\documentclass[11pt,a4paper]{article}
\usepackage{times,latexsym}
\usepackage{url}
\usepackage[T1]{fontenc}
\usepackage{amssymb}
\usepackage{amsmath}
\usepackage{cuted}
\usepackage{graphicx}
\usepackage{appendix}
\usepackage{scrextend}
\usepackage{xcolor}
\usepackage{multirow}
\usepackage[acceptedWithA]{tacl2021v1}
\usepackage{tacl2021v1}

\usepackage{xspace,mfirstuc,tabulary}

\newif\iftaclinstructions
\taclinstructionsfalse 
\iftaclinstructions
\renewcommand{\confidential}{}
\renewcommand{\anonsubtext}{(No author info supplied here, for consistency with
TACL-submission anonymization requirements)}
\newcommand{\instr}
\fi

\iftaclpubformat 

\else

\fi

\title{A Survey in Mathematical Language Processing}

 \author{
   Jordan Meadows$^1$
   \and
   Andr\'{e} Freitas$^{1,2}$
   \\
   \
   $^1$Department of Computer Science, University of Manchester, UK
   \\
   $^2$Idiap Research Institute, Switzerland
   \\
   \texttt{jordan.meadows@postgrad.manchester.ac.uk}
   \\
   \texttt{andre.freitas@idiap.ch}
   \
 }

\date{}

\begin{document}
\maketitle
\begin{abstract}

Automating discovery in mathematics and science will require sophisticated methods of information extraction and abstract reasoning, including models that can convincingly process relationships between mathematical elements and natural language, to produce problem solutions of real-world value. We analyze mathematical language processing methods across five strategic sub-areas (identifier-definition extraction, formula retrieval, natural language premise selection, math word problem solving, and informal theorem proving) in recent years, highlighting prevailing methodologies, existing limitations, overarching trends, and promising avenues for future research. 

\end{abstract}

\section{Introduction}

\begin{addmargin}[1em]{2em}
\textit{Prove that there is no function $f$ from the set of non-negative integers into itself
such that $f(f(n)) = n + 1987$ for every $n$.}\newline
\end{addmargin}

\begin{addmargin}[1em]{2em}
\textit{Show that the nearest neighbour interaction Hamiltonian of an electronic quasiparticle in Graphene can be written as $\mathcal{H} = \hbar\Omega\sum_{\textbf{q}}(f_{\textbf{q}}b^{\dagger}_{\textbf{q}}a_{\textbf{q}} + f^*_{\textbf{q}}a^{\dagger}_{\textbf{q}}b_{\textbf{q}})$}.\newline
\end{addmargin}

\begin{addmargin}[1em]{2em}
\textit{How is the sun's atmosphere hotter than its surface?}\newline
\end{addmargin}

\noindent If we hope to use machines to derive \textit{mathematically rigorous} and explainable solutions to address such questions, models must reason over both natural language and mathematical elements such as equations, expressions, and variables. Given some input problem description, the ideal model is at least capable of recalling relevant statements \textit{(premise selection)}, assigning contextual descriptions to math elements within that text \textit{(identifier-definition extraction)}, and performing robust manipulation of equations and expressions towards an explainable reasoning argument \textit{(informal theorem proving)}. Previous years have advanced many of the components required to deliver this vision. Transformer-based~\cite{vaswani2017attention} large language models (LLMs)~\cite{brown2020language, chen2021evaluating} have begun to exhibit mathematical~\cite{rabe2020mathematical} and logical~\cite{clark2020transformers} capabilities. Graph-based models also show competence in premise selection~\cite{ferreira2020premise}, math question answering~\cite{feng2021graphmr}, and math word problems (MWPs)~\cite{zhang2022hgen}. The evolutionary path of mathematical language processing can be traced from MWPs~\cite{feigenbaum1963computers, bobrow1964natural, charniak1969computer} and linguistic analysis of formal proofs~\cite{zinn1999understanding,zinn2003computational}, to the present day, where transformers and graph-based models deliver leading metrics in math and language reasoning tasks, complemented by symbolic methods~\cite{zhong2022evaluating}. This survey provides a synthesis of this recent evolutionary arch: We consider five representative tasks with examples, describe contributions leading to the current state-of-the-art, discuss notable limitations of the current solutions, overarching trends, and promising research directions. 

\begin{figure*}[htp!]
    \centering
    \includegraphics[width=1\textwidth]{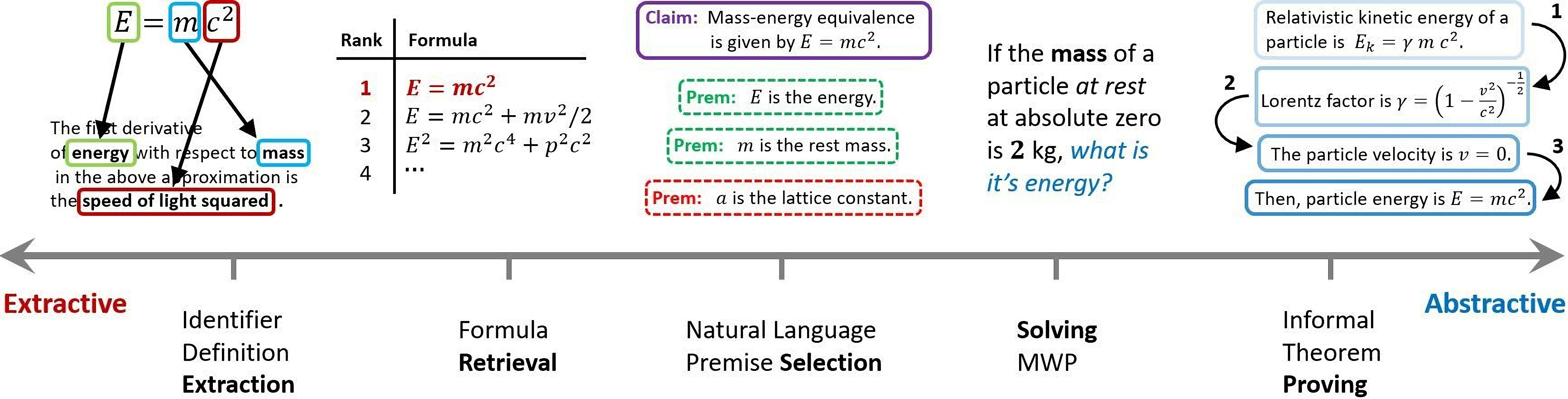}
    \caption{\textit{Extractive} tasks tend to not require inference chains to solve them, compared to more \textit{abstractive} tasks. \textit{Identifier-definition extraction} assigns identifiers $(\textit{e.g., } \psi(x))$ to their context. \textit{Formula retrieval} considers the structure of formulae, and scores them based on similarity to a query formula. \textit{Premise selection} selects statements most likely to be useful for solving a proof. \textit{Solving MWPs} (math word problems) involves calculating solutions to arithmetic problems. \textit{Informal theorem proving} involves the production of proofs and inference chains combining natural and mathematical language.}
    \label{fig:scale}
\end{figure*}

\section{Representative Tasks}

There is an abundance of tasks considering mathematical language, such as question answering~\cite{hopkins2019semeval,feng2021graphmr,lewkowycz2022solving, mansouri2022overview} and headline generation~\cite{yuan2020automatic, peng2021mathbert}. \textit{Mathematical language processing} (MLP) itself has been described in the context of various targeted texts, such as linking variables to descriptions~\cite{pagael2014mathematical}, grading answers~\cite{lan2015mathematical}, and deriving abstract representations for downstream applications~\cite{wang2021mathematical}. We take an inclusive stance, selecting a few choice tasks spanning surface-level retrieval, as seen in \textit{identifier-definition extraction} and \textit{formula retrieval} tasks, through models which require the encoding of formal abstractions and  implicit reasoning chains, such as \textit{solving MWPs} and \textit{informal theorem proving}. These areas are projected onto an inference spectrum displayed in Fig.~\ref{fig:scale}. Extractive tasks are positioned to the surface form of the text (information retrieval perspective), including identification of relevant mathematical statements, ranking lists of formulae, and linking variables to contextual definitions. Logical puzzle solvers~\cite{groza2022question} and informal reasoning generation models~\cite{lewkowycz2022solving} exist far into the abstractive side, due to the \textit{step-wise} and sometimes symbolic reasoning required to address them. The use of ``formal'' versus ``informal'' differentiates strict automated theorem prover (ATP) approaches requiring the use of a consistent formal language representation~\cite{rudnicki1992overview} and hard-coded logic~\cite{bansal2019holist}, from approaches that input mathematical language and infer without necessary reliance on strict symbolic and logical inference mechanisms. Autoformalization~\cite{szegedy2020promising,wuautoformalization2022} aims to cross this divide. We consider informal methods for solving five representative tasks in this context, with examples given below, visually displayed in Fig.~\ref{fig:scale}. \newline \noindent \textbf{Identifier-definition extraction}. The assignment of meaning to otherwise vague mathematical elements. Without context, equations such as $\mathbf{p} = \hbar \mathbf{k}$ are ambiguous. \textit{What meaning is attributed to} $\mathbf{k}$?  This task involves finding (identifier, definition) pairs, such as $(\mathbf{k}, \text{wavevector})$~\cite{kristianto2012extracting,stathopoulos2018variable}. \newline \noindent \textbf{Formula retrieval.} Mathematical language includes math elements written in markup languages such as LaTeX. Given a query formula, the Wikipedia Formula Browsing task~\cite{zanibbi2016ntcir,mansouri2022overview} involves ranking a list of candidate formulae in terms of their similarity to that formula. For example, given the query $x^2 + y^2 = r^2$, the formula $a^2 + b^2 = c^2$ should rank higher than $y = mx + c$. \newline \noindent \textbf{Natural language premise selection (NLPS).} Given a mathematical statement $s$ that requires proof, and a collection of premises $P$, this task consists of retrieving the premises in $P$ that are most likely to be useful
for proving $s$~\cite{ferreira2020natural,valentino-etal-2022-textgraphs}. For example, given the purple claim statement in Fig.~\ref{fig:scale}, a NLPS model should select the green statements as premises, excluding the red. \newline \noindent \textbf{Math word problem solving.} Solving arithmetic~\cite{roy2016solving} or algebra~\cite{kushman2014learning} word problems. \textit{Andrew has $3$ dogs. If they each give birth to $2$ others, how many dogs will he have?} An example requiring premise selection and identifier-definition extraction is given in Fig.~\ref{fig:scale}. \newline \noindent \textbf{Informal theorem proving.} Outputting reasoning chains from premises in order to ``prove" a mathematical language statement. From Fig.~\ref{fig:scale}, the energy of the particle is $E_k = \gamma m c^2$. Substituting $v = 0$ into the Lorentz factor gives $\gamma = 1$, and substituting $\gamma = 1$ into $E_k = \gamma m c^2$ gives $E_k = mc^2$. Such informal reasoning does not rely on formal frameworks, such as Fitch-style proofs, to infer quantitative results~\cite{lewkowycz2022solving}.

\section{Methods}

We highlight key points abstracted from task approaches in bold, give an overview of methods in Table 1, and discuss approach specific limitations in the appendix.

\begin{table*}[htp!]
\centering
\resizebox{1.03\textwidth}{!}{%
\begin{tabular}{llllll}
\hline
\multicolumn{1}{c}{\textbf{Work}} &
  \multicolumn{1}{c}{\textbf{Learning}} &
  \multicolumn{1}{c}{\textbf{Approach}} &
  \multicolumn{1}{c}{\textbf{Dataset}} &
  \multicolumn{1}{c}{\textbf{Metrics}} &
  \multicolumn{1}{c}{\textbf{Math Format}}
  \\ \hline
\multicolumn{6}{l}{\textbf{Identifier-Definition Extr.}} \\

\citet{kristianto2012extracting} & S & CRF with linguistic pattern features & arXiv papers & P, R, F1 & MathML\\

\citet{kristianto2014extracting} & S & SVM with linguistic pattern features & arXiv papers & P, R, F1 & MathML\\

\citet{pagael2014mathematical} & R & Gaussian heuristic ranking & Wikipedia articles & P@K, R@K & MathML\\

\citet{schubotz2016semantification} & UNS & Gaussian ranking + K-means clusters & NTCIR-11 & P, R, F1 & LaTeX\\

\citet{schubotz2017evaluating} & S & Gaussian rank + pattern matching + SVM & NTCIR-11 & P, R, F1 & LaTeX\\

\citet{stathopoulos2018variable} & S & Link prediction with BiLSTM & arXiv papers & P, R, F1 & MathML\\

\citet{alexeeva2020mathalign} & R & Odin grammar and open-domain causal IE & MathAlign-Eval & P, R, F1 & LaTeX\\

\citet{jo2021modeling} & S & BERT fine-tuning & S2ORC & Top1, Top5, MRR & LaTeX\\

\citet{ferreira-etal-2022-integer} & S & SciBERT fine-tuning with aug. data & arXiv papers & P, R, F1 & MathML\\

\citet{van2022machamp} & S & Shared encoder + multi-task decoders & Symlink & P, R, F1, F-score & LaTeX\\

\citet{ping2022team} & S & BERT fine-tuning with aug. data & Symlink & P, R, F1, F-score & LaTeX\\ 

\citet{popovic2022aifb} & S & SciBERT enc. with entity and relation extr. & Symlink & P, R, F1, F-score & LaTeX\\

\citet{lee2022jbnu} & S & SciBERT enc. with MRC + tokenizer & Symlink & P, R, F1, F-score & LaTeX\\

\hline

\multicolumn{6}{l}{\textbf{Formula Retrieval}}\\

\citet{kristianto2014mcat} & S + R & SVM desc. extr. + leaf-root path search & NTCIR-11 & P@5, P@10, MAP & MathML\\

\citet{kristianto2016mcat} & S + R & MCAT (2014) + multiple linear regr. & NTCIR-12 & P@K & MathML\\

\citet{zanibbi2016multi} & R & Inverted index rank + MSS rerank search & NTCIR-11 & R@K, MRR & MathML\\

\citet{davila2017layout} & S & \begin{tabular}[c]{@{}l@{}}Two-stage search for OPT and SLT\\ merged with linear regression\end{tabular} & NTCIR-12 & P@K, Bpref, nDCG@K & LaTeX\\

\citet{zhong2019structural} & R & \begin{tabular}[c]{@{}l@{}}OPT leaf-root path search\\ with K largest subexpressions\end{tabular} & NTCIR-12 & P@K, Bpref & LaTeX\\

\citet{mansouri2019tangent} & UNS & n-gram fastText OPT and SLT embs & NTCIR-12 & Bpref & LaTeX/MathML\\

\citet{peng2021mathbert} & SS & \begin{tabular}[c]{@{}l@{}}pre-training BERT with tasks related\\ to arXiv math-context pairs and OPTs\end{tabular} & NTCIR-12 & Bpref & LaTeX\\

\citet{zhong2022evaluating} & R + S & Approach0 + dense passage retrieval enc.& NTCIR-12 & Bpref & LaTeX\\ 

\hline
  \multicolumn{6}{l}{\textbf{Premise Selection}} \\

\citet{ferreira2020premise} & S &
  DGCNN for link prediction &
  PS-ProofWiki &
  P, R, F1 &
  LaTeX\\
  
\citet{ferreira2021star} & S &
  \begin{tabular}[c]{@{}l@{}}Self-attention for math and language\\ + BiLSTM with siamese network\end{tabular} &
  PS-ProofWiki &
  P, R, F1 &
  LaTeX\\
  
  \citet{coavoux2021learning} &
  S &
  Weighted bipartite matching + self-attention &
  SPM &
  MRR, Acc &
  MathML\\
  
\citet{hancontrastive} & S & LLM fine-tuning (webtext + webmath) & NaturalProofs & R@K, avgP@K, full@K & LaTeX\\

\citet{welleck2021naturalproofs} & S & Fine-tuning BERT with pair/joint param. & NaturalProofs & MAP, R@K, full@K & LaTeX\\

\citet{dastgheib2022keyword} & UNS & keywords fastText emb. with Jacardian sim & PS-ProofWiki & MAP & LaTeX\\

\citet{kovriguina2022textgraphs} & SS & MathBERT enc. with GPT-3 prompting & PS-ProofWiki & MAP & LaTeX\\

\citet{kadusabe2022snlp} & SS & SMPNet with cosine similarity & PS-ProofWiki & MAP & LaTeX\\

\citet{tran2022ijs} & SS & RoBERTa with Manhattan similarity & PS-ProofWiki & MAP & LaTeX\\

\hline
\multicolumn{6}{l}{\textbf{MWP Solving}}\\

\citet{liu2019tree} & S & BiLSTM seq enc. + LSTM tree-based dec. & Math23K & Acc & NL\\

\citet{xie2019goal} & S & GRU encoder + GTS decoder & Math23K & Acc & NL\\

\citet{li2020graph} & S & \begin{tabular}[c]{@{}l@{}}word-word graph + phrase structure graph\\ Hetero. graph enc. + LSTM tree-based dec. \end{tabular}  & MAWPS, MATHQA & Acc & NL\\

\citet{zhang2020graph} & S & \begin{tabular}[c]{@{}l@{}}Word-number graph enc. +\\ Number-comp graph enc. + GTS dec.\end{tabular} & MAWPS, Math23K & Acc & NL\\

\citet{shen2020solving} & S & Seq multi-enc. + tree-based multi-dec. & Math23K & Acc & NL\\

\citet{kim2020point} & S & ALBERT seq enc. + Transformer seq dec. & \begin{tabular}[c]{@{}l@{}}ALG514, DRAW-1K, \\ MAWPS\end{tabular} & Acc & NL\\

\citet{qin2020semantically} & S & \begin{tabular}[c]{@{}l@{}}Bi-GRU seq enc. + \\ semantically-aligned GTS-based dec.\end{tabular} & \begin{tabular}[c]{@{}l@{}}HMWP, ALG514,\\ Math23K, Dolphin18K\end{tabular} & Acc & NL\\

\citet{cao2021bottom} & S & GRU-based encoder + DAG-LSTM decoder & DRAW-1K, Math23K & Acc & NL\\

\citet{lin2021hms} & S & Hierarchical GRU seq encoder + GTS decoder & Math23K, MAWPS & Acc & NL\\

\citet{qin2021neural} & S & Bi-GRU enc. + GTS dec. with att. and UET & Math23K, CM17K & Acc & NL\\

\citet{liang2021mwp} & S & BERT encoder + GTS decoder & Math23K, APE210K & Acc & NL\\

\citet{zhang2022hgen} & S & \begin{tabular}[c]{@{}l@{}} word-word + word-num + num-comp graph\\ Heterogeneous graph encoder + GTS decoder\end{tabular} & MAWPS, Math23K & Acc & NL\\

\citet{jie2022learning} & S & RoBERTa enc. with bottom-up relation extr. & \begin{tabular}[c]{@{}l@{}} MAWPS, Math23K,\\ SVAMP, MathQA\end{tabular} & Acc & NL\\

\citet{zhang2022multi} & S & \begin{tabular}[c]{@{}l@{}} Top-down and bottom-up reasoning with\\ knowledge injection and contrastive learning\end{tabular} &  \begin{tabular}[c]{@{}l@{}} MAWPS, Math23K,\\ MathQA\end{tabular}& Acc & NL\\

\hline
\multicolumn{6}{l}{\textbf{Informal Theorem Proving}} \\

\citet{wang2020exploration} & S + UNS & RNNs, LSTMs, transformers & \begin{tabular}[c]{@{}l@{}}LaTeX, Mizar,\\  TPTP, ProofWiki\end{tabular} & \begin{tabular}[c]{@{}l@{}}BLEU, Perplexity\\ Edit distance\end{tabular} & LaTeX\\

\citet{welleck2021naturalproofs} & S & Fine-tuning BERT with pair/joint param. & NaturalProofs & MAP & LaTeX\\

\citet{welleck2021towards} & SS & \begin{tabular}[c]{@{}l@{}} BART enc. with denoising pre-training\\ and Fusion-in-Decoder\end{tabular} & NaturalProofs & SBleu, Meteor, Edit, P, R, F1 & LaTeX\\

\citet{wuautoformalization2022} & S & Fine-tuned LLMs + formal theorem prover & MiniF2F, MATH & Acc & LaTeX/Isabelle\\

\citet{lewkowycz2022solving} & SS/S & Fine-tuned PaLM model & \begin{tabular}[c]{@{}l@{}} MATH, GSM8K,\\ MMLU-STEM\end{tabular} & Acc & LaTeX/NL\\

\end{tabular}%
}
\caption{
Summary of different approaches for addressing tasks related to mathematical language processing. The methods are categorised in terms of (i) Learning: Supervised (S), Self-supervised (SS), Unsupervised (UNS), Rule-based (R) (no learning); (iii) Approach; (iv) Dataset; (v) Metrics: MAP (Mean Average Precision), P@K (Precision at K), Perplexity, P (Precision), R (Recall), F1, Acc (Accuracy), BLEU, METEOR, MRR (Mean Reciprocal Rank), Edit (edit distance); (vi) Math format: MathML, LaTeX, natural language (NL), Isabelle formal language. Diagrammatic representations of approaches in identifier-definition extraction (Fig.~\ref{fig:IDE_taxonomy}), formula retrieval (Fig.~\ref{fig:FR_taxonomy}), and MWP solving (Fig.~\ref{fig:MWP_taxonomy}) can be found in the Appendix.}
\label{tab:related_work:works_for_maths_processing}
\end{table*}

\subsection{Identifier-Definition Extraction} 

A significant proportion of variables or identifiers in formulae or text are explicitly defined within a discourse context~\cite{wolska2010symbol}. Descriptions are usually local to the first instance of the identifiers in the discourse. It is the broad goal of identifier-definition extraction and related tasks to pair-up variables with their intended meaning. 
\paragraph{The task has not converged to a canonical form.} Despite the clarity of its overall aim, the task has materialised into different forms: \citet{kristianto2012extracting} predict descriptions given \textit{expressions}, \citet{pagael2014mathematical} predict descriptions given identifiers through \textit{identifier-definition extraction}, \citet{stathopoulos2018variable} predict if a type matches a variable through \textit{variable typing}, and
\citet{jo2021modeling} predict notation given context through \textit{notation auto-suggestion} and \textit{notation consistency checking} tasks. More concretely, \textit{identifier-definition extraction}~\cite{schubotz2016semantification} involves \textit{scoring} identifier-definiens pairs, where a definiens is a potential natural language description of the identifier. Given graph nodes from predefined variables $V$ and types $T$, \textit{variable typing}~\cite{stathopoulos2018variable} is the task of \textit{classifying} whether edges $V \times T$ are either existent (positive) or non-existent (negative), where a positive classification means a variable matches with the type. \textit{Notation auto-suggestion}~\cite{jo2021modeling} uses the text of both the sentence containing notation and the previous sentence to \textit{model} future notation from the vocabulary of the tokenizer. This area can be traced from an early ranking task~\cite{pagael2014mathematical} reliant on heuristics and rules~\cite{alexeeva2020mathalign}, through ML-based edge classification~\cite{stathopoulos2018variable}, to language modelling with Transformers~\cite{jo2021modeling}. Different datasets are proposed for each task variant. \paragraph{There is a high variability in scoping definitions.} The scope from which identifiers are linked to descriptions varies significantly, and it is difficult to compare model performance even when tackling the same variant of the task~\cite{schubotz2017evaluating,alexeeva2020mathalign}. At a local context, models such as \citet{pagael2014mathematical} and \citet{alexeeva2020mathalign} match identifiers with definitions from the same document ``as the author intended", while other identifier-definition extraction methods~\cite{schubotz2016semantification,schubotz2017evaluating} rely on data external to a given document, such as links to semantic concepts on Wikidata and NTCIR-11 test data~\cite{schubotz2015challenges}. At a broader context, the variable typing model proposed in \citet{stathopoulos2018variable} relies on an external dictionary of types~\cite{stathopoulos2015retrieval,stathopoulos2016mathematical,stathopoulos2018variable} extracted from both the Encyclopedia of Mathematics\footnote{https://encyclopediaofmath.org} and Wikipedia.
\paragraph{Vector representations have evolved to transfer knowledge from previous tasks, allowing downstream variable typing tasks to benefit from pre-trained embeddings.} Overall, vector representations of text have evolved from feature-based vectors learned from scratch for a single purpose, to the modern paradigm of pre-trained embeddings re-purposed for novel tasks. \citet{kristianto2012extracting} input pattern features into a conditional random fields model for the purpose of identifying definitions of expressions in LaTeX papers while \citet{kristianto2014extracting} learn vectors through a linear-kernel SVM with input features comprising of sentence patterns, part-of-speech (POS) tags, and tree structures. \citet{stathopoulos2018variable} extend this approach by adding type- and variable-centric features as a baseline also with a linear kernel. Alternatively, \citet{schubotz2017evaluating} use a Gaussian scoring function~\cite{schubotz2016getting} and pattern matching features~\cite{pagael2014mathematical} as input to an SVM with a radial basis function (RBF) kernel, to account for non-linear feature characteristics. Alternative classification methods~\cite{kristianto2012extracting, stathopoulos2018variable} do not use input features derived from non-linear functions, such as the Gaussian scoring function, and hence use linear kernels. Embedding spaces have been learned in this context for the purpose of \textit{ranking} identifier-definiens pairs through latent semantic analysis at the document level, followed by the application of clustering techniques and methods of relating clusters to namespaces inherited from software engineering~\cite{schubotz2016semantification}. These cluster-based namespaces are later used for \textit{classification}~\cite{schubotz2017evaluating} rather than ranking, but do not positively impact SVM model performance, despite previous evidence suggesting they resolve co-references~\cite{duval2002metadata} such as ``$E$ is energy" and ``$E$ is expectation value". Neither clustering nor namespaces have been further explored in this context. More recent work learns context-specific word representations after feeding less specific pre-trained word2vec~\cite{mikolov2013efficient,stathopoulos2016mathematical} embeddings to a bidirectional LSTM for classification~\cite{stathopoulos2018variable}. The most recent work predictably relies on more sophisticated pre-trained BERT embeddings~\cite{devlin2018bert} for the language modelling of mathematical notation~\cite{jo2021modeling}. VarSlot~\cite{ferreira-etal-2022-integer} obtains SOTA results on variable typing~\cite{stathopoulos2018variable}, and demonstrates robustness to variable renaming, by fine-tuning the sentence transformers~\cite{reimers2019sentence} SciBERT~\cite{beltagy2019scibert} encoder on \textit{augmented data}, learning separate representation spaces for variables and mathematical language statements. Four BERT encoder-based approaches~\cite{lee2022jbnu,popovic2022aifb,ping2022team, van2022machamp} were submitted to the Symlink task~\cite{lai2022semeval}, following the trend of knowledge transfer through pretrained embeddings.

\subsection{Formula Retrieval}

\begin{figure}[h!]
    \centering
    \includegraphics[width=0.45\textwidth]{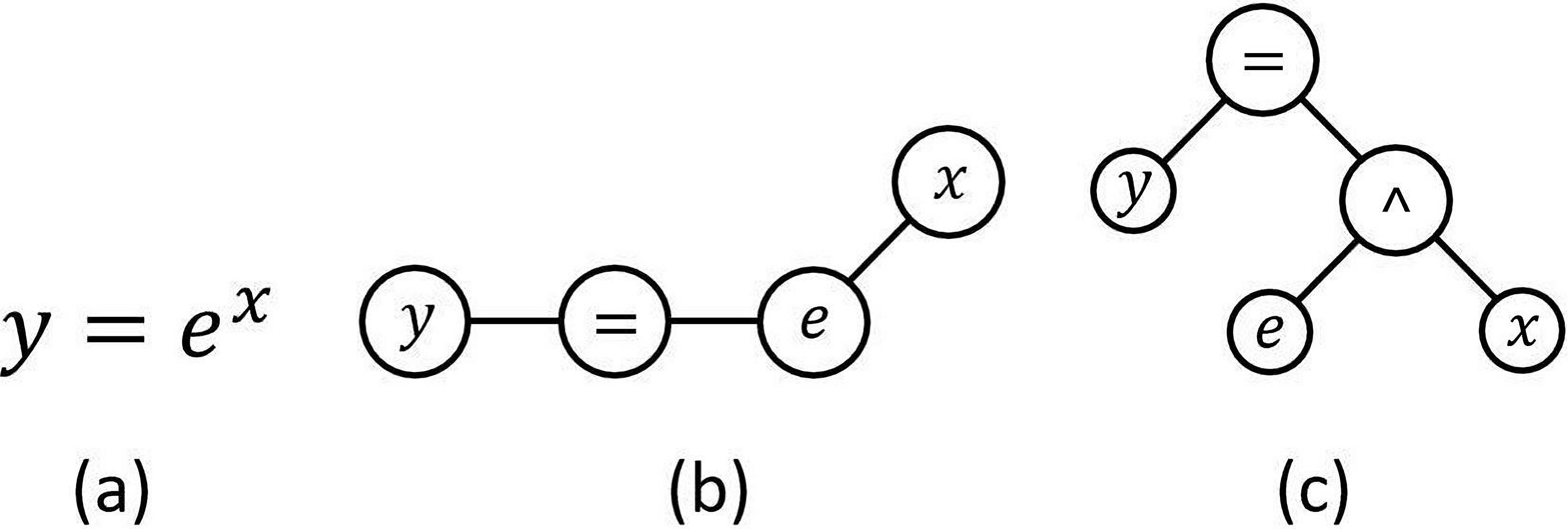}
    \caption{Formula (a) $y=e^x$ with its Symbol Layout Tree
    (SLT) (b), and Operator Tree (OPT) (c). SLTs represent formula appearance by the spatial arrangements of math symbols, while OPTs
    define the mathematical operations represented in expressions. For more detail, see \citet{mansouri2019tangent}.}
    \label{tree_repr}
\end{figure}

The task of retrieving similar equations to a query equation, with applications in math-aware search engines~\cite{mansouri2022advancing}. \citet{guidi2016survey,Zanibbi_2011} emphasise the encoding of formulae and their context for retrieval tasks. 
\paragraph{Combining formula tree representations improves retrieval.} There are two prevalent types of tree representations of formulae: Symbol Layout Trees (SLTs) and Operator Trees (OPTs), shown in Fig.~\ref{tree_repr}. 

\noindent Methods reliant solely on SLTs, such as the early versions of the Tangent retrieval system~\cite{pattaniyil2014combining,zanibbi2015tangent,zanibbi2016multi}, or solely OPTs~\cite{zhong2019structural,zhong2020accelerating} tend to return less relevant formulae from queries. OPTs capture formula semantics while SLTs capture visual structure~\cite{mansouri2019tangent}. Effective representation of both formula layout and semantics within a single vector allows a model exploit both representations. Tangent-S~\cite{davila2017layout} was the first evolution of the Tangent system to outperform the NTCIR-11~\cite{aizawa2014ntcir} overall best performer, MCAT~\cite{kristianto2014mcat,kristianto2016mcat}, which encoded path and sibling information from MathML Presentation (SLT-based) and Content (OPT-based). Tangent-S jointly integrated SLTs and OPTs by combining scores for each representation through a simple linear regressor. Later, Tangent-CFT~\cite{mansouri2019tangent} considered SLTs and OPTs through a fastText~\cite{bojanowski2017enriching} n-gram embedding model using tree tuples. MathBERT~\cite{peng2021mathbert} does \textit{not} explicitly account for SLTs, claiming that LaTeX markup somewhat accounts for SLTs, and therefore encode OPTs. They pre-train the BERT~\cite{devlin2018bert} model with targeted objectives each accounting for different aspects of mathematical text. They account for OPTs by concatenating node sequences to formula + context BERT input sequences, and by formulating OPT-based structure-aware pre-training tasks learned in conjunction with masked language modelling (MLM). 
\paragraph{Leaf-root path tuples deliver an effective mechanism for embedding relations between symbol pairs.} Leaf-root path tuples are now ubiquitous in formula retrieval~\cite{zanibbi2015tangent,zanibbi2016multi,davila2017layout,zhong2019structural,mansouri2019tangent,zhong2020accelerating} and their use for NTCIR-11/12 retrieval has varied since their conception~\cite{stalnaker2015math}. Initially~\cite{pattaniyil2014combining} pair tuples were used within a TF-IDF weighting scheme, then \citet{zanibbi2015tangent,zanibbi2016multi} proposed an appearance-based similarity metric using SLTs, \textit{maximum subtree similarity} (MSS). OPT tuples are integrated~\cite{davila2017layout} later on. \citet{mansouri2019tangent} treat tree tuples as words, extract n-grams, and learn fastText~\cite{bojanowski2017enriching} formula embeddings. \citet{zhong2019structural,zhong2020accelerating} forgo machine learning altogether with an OPT-based heuristic search (Approach0) through a generalisation of MSS~\cite{zanibbi2016multi}. Leaf-root path tuples effectively map symbol-pair relations and account for formula substructure, but there is dispute on how best to integrate them into existing machine learning or explicit retrieval frameworks. 
\paragraph{Purely explicit methods still deliver competitive results.} Explicit representation methods are those that rely on prescribed representations (structural relations and associated types) rather than learned implicit relationships. Tangent-CFT~\cite{mansouri2019tangent} and MathBERT~\cite{peng2021mathbert} are two models to employ learning techniques beyond the level of linear regression. Each model is integrated with Approach0~\cite{zhong2019structural} through the linear combination of individual model scores. This respectively forms the TanApp and MathApp baselines in \citet{peng2021mathbert}. Approach0 achieves the highest full bpref score of the individual models. While we focus primarily on the NTCIR-12 dataset, recent work~\cite{zhong2022evaluating} evaluates a selection of transformer-based models on both NTCIR-12 and ARQMath-2~\cite{mansouri2021overview} datasets. They confirm MathBERT delivers SOTA performance on partial bpref, and Approach0 combined with a fine-tuned dense passage retrieval (DPR) model~\cite{karpukhin2020dense} outperforms on full bpref (Approach0 + DPR). Combining explicit similarity-based search~\cite{zhong2019structural, meadows2021similarity} with modern encoders~\cite{khattab2020colbert, karpukhin2020dense} delivers leading performance.

 

\subsection{Natural Language Premise Selection}

Formal and informal premise selection both involve \textit{the selection of relevant statements for proving a given conjecture}~\cite{irving2016deepmath,wang2017premise,ferreira2020natural}. The difference lies in the language in which the premises and related proof elements are encoded (either conforming to a logical form or as they appear in mathematical text). Mathematical language as it occurs in papers and textbooks~\cite{wolska2004analysis} is not compatible with existing provers without \textit{autoformalization}; a widely acknowledged bottleneck for the construction of formal proof libraries~\cite{irving2016deepmath}. Typically, when reasoning over large formal libraries comprising thousands of premises, the performance of ATPs degrades considerably, while for a given proof only a fraction of the premises are required to complete it~\cite{urban2010evaluation,alama2014premise}. Theorem proving is essentially a search problem with a combinatorial search space, and the goal of \textit{formal} premise selection is to reduce the space, making theorem proving tractable~\cite{wang2017premise}. While formal premises are written in the languages of formal libraries such as Mizar~\cite{rudnicki1992overview}, \textit{informal} premises, as seen in ProofWiki\footnote{https://proofwiki.org/wiki/Main\_Page}, are written in combinations of natural language and LaTeX~\cite{ferreira2020natural,welleck2021naturalproofs}. Proposed approaches either rank~\cite{hancontrastive} or classify~\cite{ferreira2020premise,ferreira2021star} candidate premises for a given proof. \textit{Natural language premise selection} was originally formulated as pairwise relevance classification, evaluated with $F_1$~\cite{ferreira2020premise, ferreira2021star}, but has since been evaluated with ranking metrics~\cite{valentino-etal-2022-textgraphs}. Alternatively, \citet{welleck2021naturalproofs} propose \textit{mathematical reference retrieval} as an analogue of premise selection. The goal is to retrieve the set of references (theorems, lemmas, definitions) that occur in its proof, formulated as a \textit{ranking} problem. 
\paragraph{Separate mechanisms for representing mathematics and natural language can improve performance.} Regardless of the task variation, most current methods do not fully discriminate the semantics of mathematics and natural language, not specifically accounting for aspects of each modality. \citet{ferreira2020premise} extract a dependency graph representing dual-modality mathematical statements as nodes, and solve a link prediction task~\cite{zhang2018link}. Recent transformer baselines \cite{ferreira2020premise,welleck2021naturalproofs,hancontrastive,coavoux2021learning}, and those at the shared NLPS task~\cite{valentino-etal-2022-textgraphs}, also do not differentiate between mathematical elements and natural language~\cite{tran2022ijs, kadusabe2022snlp, kovriguina2022textgraphs}. STAR~\cite{ferreira2021star} purposefully separates the two modalities, encoding distinct representations through self-attention. Explicit disentanglement of the modalities forces STAR to exploit relationships between natural language and mathematics, through the BiLSTM layer. Neuroscience research suggests the brain handles mathematics separately to language~\cite{butterworth2002mathematics, amalric2016origins, kulasingham2021cortical}.

\subsection{Math Word Problems}

\noindent Solving math word problems dates back to the dawn of artificial intelligence research~\cite{feigenbaum1963computers,bobrow1964natural,charniak1969computer}. It can be defined as the task of translating a problem description paragraph into a set of equations to be solved~\cite{li2020graph}. We focus on trends in the task since 2019, as a detailed survey~\cite{zhang2019gap} captures prior work. 
\paragraph{Use of dependency graphs is instrumental to support inference.} In graph-based approaches to solving MWPs, embeddings of words, numbers, or relationship graph nodes, are learned through \textit{graph encoders} which feed information through to tree (or sequence) decoders. Embeddings are decoded into expression trees which determine the problem solution. \citet{li2020graph} learn the mapping between a heterogeneous graph representing the input problem, and an output tree. The graph is constructed from word nodes with relationship nodes of a parsing tree. This is either a dependency parse tree or constituency tree. \citet{zhang2020graph} represent \textit{two} separate graphs: a \textit{quantity cell graph} associating descriptive words with problem quantities, and a \textit{quantity comparison graph} which retains numerical qualities of the quantity, and leverages heuristics to represent relationships between quantities such that solution expressions reflect a more realistic arithmetic order. \citet{shen2020solving} also extract \textit{two} graphs: a dependency parse tree and numerical comparison graph. \citet{zhang2022hgen} construct a heterogeneous graph from three subgraphs: a \textit{word-word graph} containing syntactic and semantic relationships between words, a \textit{number-word graph}, and a \textit{number comparison graph}. Although other important differences exist (such as decoder choice), it seems models benefit from relating linguistic aspects of problem text through separate graphs.
\paragraph{Multi-encoders and multi-decoders improve performance by combining complementary representations.} Another impactful decision is the choice of encoder/decoder, and whether to consider alternative representations of a problem. To highlight this, we consider the following comparison. \citet{shen2020solving} and \citet{zhang2020graph} each extract two graphs from the problem text. One is a number comparison graph, and the other relates word-word pairs~\cite{shen2020solving} or word-number pairs~\cite{zhang2020graph}. They both encode \textit{two} graphs rather than one heterogeneous graph~\cite{li2020graph,zhang2022hgen}. They both use a similar tree-based decoder~\cite{xie2019goal}. A key difference is that \citet{shen2020solving} includes \textit{an additional sequence-based encoder and decoder}. The sequence-based encoder first obtains a textual representation of the input paragraph, then the graph-based encoder integrates the two encoded graphs. Then tree-based and sequence-based decoders generate \textit{different equation expressions} for the problem with an additional mechanism for optimising solution expression selection. In their own work, \citet{shen2020solving} demonstrate the impact of multi-encoders/decoders over each encoder/decoder option individually through ablation. \citet{zhang2022multi} similarly combine top-down and bottom-up reasoning to achieve leading results.
\paragraph{Goal-driven decompositional tree-based decoders are a significant component in the state-of-the-art.} Introduced in \citet{xie2019goal}, this class of decoder is considered by most of the discussed approaches, and includes non-graph-based models~\cite{qin2021neural,liang2021mwp}. In GTS, goal vectors guide construction of expression subtrees (from token node embeddings) in a recursive manner, until a solution expression tree is generated. Proposed models do expand on the GTS-based decoder through the inclusion of semantically-aligned universal expression trees~\cite{qin2020semantically, qin2021neural}, though this adaptation is not as widely used. Some state-of-the-art~\cite{liang2021mwp,zhang2022hgen} models follow the GTS decoder closely. 
\paragraph{Language models that transfer knowledge learned from auxiliary tasks rival models based on explicit graph representation of problem text.} As an alternative to encoding explicit relations through graphs, other work~\cite{kim2020point,qin2021neural,liang2021mwp} relies on pre-trained transformer-based models, and those which incorporate auxiliary tasks assumed relevant for solving MWPs, to learn such relations latently. However, it seems the case that auxiliary tasks alone do not deliver competitive performance~\cite{qin2020semantically} without the extensive pre-training efforts with large corpora, as we see with BERT-based transformer models. These use either both the ALBERT~\cite{lan2019albert} encoder and decoder~\cite{kim2020point}, or BERT-based encoder with goal-driven tree-based decoder~\cite{liang2021mwp}. More recent work~\cite{cao2021bottom, jie2022learning, zhang2022multi} involves \textit{iterative relation extraction} frameworks
for predicting mathematical relations between numerical tokens. 

\subsection{Informal Theorem Proving} 
Formal automated theorem proving in logic is among the most abstract forms of reasoning materialised in the AI space. There are two major bottlenecks~\cite{irving2016deepmath} formal methods must overcome: (1) translating informal mathematical text into formal language (\textit{autoformalization}), and (2) a lack of strong automated reasoning methods to fill in the gaps in already formalised human-written proofs. Informal methods either tackle autoformalization directly~\cite{wang2020exploration,wuautoformalization2022}, or circumvent it through language modelling-based proof generation~\cite{welleck2021naturalproofs,welleck2021towards}, trading formal rigour and inference control for flexibility. Transformer-based models have been proposed for mathematical reasoning~\cite{polu2020generative,rabe2020mathematical,wu2021lime}. Converting \textit{informal} mathematical text into forms which are interpretable by computers~\cite{kaliszyk2015formalizing,kaliszyk2015learning,szegedy2020promising,wang2020learning,meadows2021similarity} can strategically impact the dialogue between knowledge expressed in natural text, and a large spectrum of solvers. 
\paragraph{Autoformalization could be addressed through approximate translation and exploration rather than direct machine translation.} A long-studied and challenging endeavour~\cite{zinn1999understanding, zinn2003computational}, autoformalization involves converting informal mathematical text into language interpretable by theorem provers~\cite{kaliszyk2015learning,wang2020exploration,szegedy2020promising}. \citet{kaliszyk2015learning} propose statistical learning methods for parsing ambiguous formulae over the Flyspeck formal mathematical corpus~\cite{hales2006introduction}. Using machine translation models~\cite{luong2017neural,lample2018phrase,lample2019cross}, \citet{wang2020exploration} explore dataset translation experiments between LaTeX code extracted from ProofWiki, and formal libraries Mizar~\cite{rudnicki1992overview} and TPTP~\cite{sutcliffe1998tptp}. The supervised RNN-based neural machine translation model~\cite{luong2017neural} outperforms the transformer-based~\cite{lample2018phrase} and MLM pre-trained transformer-based~\cite{lample2019cross} models, with the performance boost stemming from its use of alignment data. \citet{szegedy2020promising} advises against such direct translation efforts, instead proposing a combination of exploration and approximate translation through predicting formula embeddings. In seq2seq models, embeddings are typically granular, encoding word-level or symbol-level~\cite{jo2021modeling} tokens. The method consists of learning mappings from natural language input to premise statements nearby the desired statement in the embedding space, traversing the space between statements using a suitable prover~\cite{bansal2019holist}. Guided mathematical exploration for real-world proofs is still an unaddressed problem and does not scale well with step-distance between current and desired conjecture. \citet{wuautoformalization2022} directly autoformalize small competition problems to Isabelle statements using language models. Similar to previous indication~\cite{szegedy2020promising}, they also autoformalize statements as targets for proof search with a neural theorem prover.
\paragraph{Need for developing robust interactive natural language theorem provers.} We discuss the closest equivalent to formal theorem proving in an informal setting. \citet{welleck2021naturalproofs} propose a \textit{mathematical reference generation} task. Given a mathematical claim, the order and number of references within a proof are predicted. A reference is a theorem, definition, or a page that is linked to within the contents of a statement or proof. Each theorem $\mathbf{x}$ has a proof containing a sequence of references $\mathbf{y} = (\mathbf{r_1}, ..., \mathbf{r_{|y|}})$, for references $\mathbf{r_m} \in \mathcal{R}$. Where the \textit{retrieval} task assigns a score to each reference in $\mathcal{R}$, the \textit{generation} task produces a variable length of sequence of references $(\mathbf{\hat{r}_1}, ..., \mathbf{\hat{r}_{|y|}})$ with the goal of matching $\mathbf{y}$, for which a BERT-based model is employed and fine-tuned on various data sources. \citet{welleck2021towards} expand on their proof generation work, proposing two related tasks: \textit{next-step suggestion}, where a step from a proof $\mathbf{y}$ (as described above) is defined as a sequence of tokens to be generated, given the previous steps and $\mathbf{x}$; and \textit{full-proof generation} which extends this to generate the full proof. They employ BART~\cite{lewis2019bart}, an encoder-decoder model pre-trained with \textit{denoising} tasks, and augment the model with reference knowledge using Fusion-in-Decoder~\cite{izacard2020leveraging}. The intermediate denoising training and knowledge-grounding improve model performance by producing better representations of (denoised) references for deployment at generation time, and by encoding reference-augmented inputs. Minerva~\cite{lewkowycz2022solving} is a language model capable of producing step-wise reasoning with mathematical language (LaTeX). They fine-tune a PaLM decoder-only model~\cite{chowdhery2022palm} on webpages containing MathJax formatted expressions, and evaluate on school-level math problems~\cite{hendrycks2021measuring,cobbe2021training}, a STEM subset of problems~\cite{hendrycks2020measuring} of varying difficulty, undergraduate-level STEM problems, and the National Math Exam in Poland. They evaluate for \textit{generalisation capabilities} by generating 20 alternative evaluation problems, perturbing problem wording and numerical values in the MATH~\cite{hendrycks2021measuring} dataset, and compare accuracy before and after the change. While they suggest ``minimal memorization'',  the numerical intervention comparison does less to support this claim.

\section{Datasets} 
 
 Various datasets have been proposed for tasks related to \textit{identifier-definition extraction} and variable typing~\cite{schubotz2016semantification,alexeeva2020mathalign,stathopoulos2018variable,jo2021modeling}, with limited adoption. The Symlink shared task~\cite{lai2022semeval} is an emerging solution, with training data, annotations of 102 papers, and high inter-annotator agreement. \textit{Formula retrieval} data exists through NTCIR-12~\cite{zanibbi2016ntcir}, which has been expanded in the most recent ARQMath task~\cite{mansouri2022overview}, removing formula duplicates and balancing query complexity. \textit{Premise selection} datasets include PS-ProofWiki~\cite{ferreira2020natural}, used in the NLPS shared task~\cite{valentino-etal-2022-textgraphs}, and NaturalProofs~\cite{welleck2021naturalproofs}. The latter is more inclusive, comprising ProofWiki, text books, and other sources. Modern consensus \textit{MWP} datasets include (easy) MAWPS~\cite{koncel2016mawps}, (medium) Math23K~\cite{wang2017deep}, and (hard) MathQA~\cite{amini2019mathqa}, comprising both Chinese and English problems. GSM8K~\cite{cobbe2021training} claims to resolve diversity, quality, and language~\cite{huang2016well} issues from previous datasets, involves step-wise reasoning and natural language solutions, with balanced difficulty. MATH~\cite{hendrycks2021measuring} is larger and more difficult than GSM8K. \textit{Informal theorem proving} data includes NaturalProofs~\cite{welleck2021naturalproofs}, and some MWP datasets involving step-wise reasoning with mathematical language, such as MATH and GSM8K. However, there is no consensus data for autoformalization or theorem proving from mathematical language input involving sequence learning. ProofNet~\cite{azerbayevproofnet} aims to remedy this, by providing 297 theorem statements expressed in both natural and formal~\cite{moura2015lean} language, at undergraduate difficulty. Some are accompanied by informal proofs. MiniF2F~\cite{zheng2021minif2f} is a neural theorem proving benchmark of Olympiad-level problems written in many formal languages. Lila~\cite{mishra2022lila} provides data for 23 math reasoning tasks. Key datasets information is described in Table~\ref{tab:datasets}.
 \paragraph{Data scarcity.} Some datasets, such as MATH and the Auxiliary Mathematics Problems and Solutions (AMPS)~\cite{hendrycks2021measuring} datasets, include detailed workings at high school to undergraduate level difficulty. If we aim to use models to produce new mathematics, equivalent datasets composed of the research workings of actual mathematicians would be invaluable. \citet{meadows2021similarity} attempt to tackle this problem for a single research paper in a very limited setting.

{\renewcommand{\arraystretch}{1.3}
 \begin{table*}[htp!]
\centering

\begin{tabular}{c|c|c}
\textbf{Name} &
\textbf{Tasks} &
\textbf{Size}
  \\ \hline
Symlink & Identifier-Def Extr. & 31K entities, 20K relations\\ \hline
ARQMath-2 Task 2 & Formula Retrieval & 100 queries, 28M formulae \\ \hline
NTCIR-12 & Formula Retrieval & 40 formula queries, 590K formulae\\ \hline
\multirow{2}{*}{PS-ProofWiki} & \multirow{2}{*}{Premise Selection} & 14K theorems, 5K definitions\\
& & 300 lemmas, 292 corollaries\\ \hline
\multirow{2}{*}{NaturalProofs} & Premise Selection & 32K theorems/proofs, 14K definitions\\
& Proof Generation & 2K corollories + axioms\\ \hline
Math23K & Math Word Problems & 23K problems\\ \hline
MAWPS & Math Word Problems & 3K problems\\ \hline
MathQA & Math Word Problems & 37K problems\\ \hline
GSM8K & Math Word Problems & 8K problems\\ \hline
\multirow{2}{*}{MATH} & Math Word Problems & \multirow{2}{*}{13K hard problems}\\
& Proof Generation & \\ \hline
ProofNet & Proof Generation & 297 theorems/proofs\\

\end{tabular}%
\caption{Key datasets for the representative tasks.}
\label{tab:datasets}
\end{table*}}

\section{Discussion}

 \textbf{State-of-the-art.} In \textit{identifier-definition extraction}, leading performance is obtained on Symlink by \citet{lee2022jbnu}, using a SciBERT encoder and MRC-based model~\cite{li2019unified}. Importantly, rather than the BERT tokenizer, they use a \textit{rule-based symbol tokenizer}, evidencing the benefits of discerning natural language from math elements. VarSlot~\cite{ferreira-etal-2022-integer} leads in variable typing, and echoes the importance of such discrimination (see Section 3.2). In \textit{formula retrieval}, SOTA methods generally include linear combinations of scores obtained from symbolic and neural models. On NTCIR-12, \citet{zhong2022evaluating} show that MathBERT leads on partial bpref, and Approach0 + DPR leads on full bpref (see Section 3.2). Approach0 + ColBERT~\cite{khattab2020colbert} leads on ARQMath-2~\cite{mansouri2021overview}. This work reinforces the importance of including formula structure across multiple tasks. In \textit{premise selection}, leading results are obtained on the shared NLPS task by a fine-tuned RoBERTa-large encoder~\cite{liu2019roberta}, computing similarity scores between statements with Manhattan distance~\cite{tran2022ijs}. However, none of the competing models discern mathematical elements from natural language, or include formula structure. In \textit{MWP solving}, the multi-view model~\cite{zhang2022multi} achieves state-of-the-art results on Math23K, MAWPS, and MathQA. Minerva, and the Diverse approach~\cite{li2022advance} based on OpenAI code-davinci-002, lead on MATH. Minerva also beats the national 57\% average by 8\% on the Polish national math exam. In \textit{informal theorem proving}, we discuss autoformalization and theorem proving from mathematical language. In the former, code-davinci-002 leads on ProofNet. In the latter, a BART-based model leads on NaturalProofs, and Codex~\cite{chen2021evaluating} fine-tuned on autoformalized theorems~\cite{wuautoformalization2022}, leads on MiniF2F. These later methods, particularly those that score highly on MATH, largely consist of fine-tuning generative LLMs also without distinctly considering mathematical content or structure.
 \paragraph{Separate representations for math and natural language.} Many models do not benefit from processing each modality separately. The leading model on Symlink uses a special tokenizer to extract math symbols from scientific documents~\cite{lee2022jbnu}. VarSlot improves variable typing by learning representation spaces for variables and mathematical language statements~\cite{ferreira-etal-2022-integer}. STAR~\cite{ferreira2021star} improves on a self-attention baseline encoding combined math/language statements, by separately encoding math and language with the same encoder. MathBERT learns embeddings from tree and latex representations of formulae, and natural language~\cite{peng2021mathbert}. The Approach0 + [encoder] models linearly combine scores from entirely different methods; one designed for formulae, and one for language~\cite{zhong2022evaluating}. Multi-view learns an embedding each for words, quantities, and operations~\cite{zhang2022multi}. All of the above are state-of-the-art and show advantage over baselines that do not invoke separate mechanisms. Despite this evidence, methods related to \textit{informal theorem proving} and \textit{premise selection}, such as Minerva, IJS~\cite{tran2022ijs}, and others, do not discriminate math from language. This is likely true for other subfields of MLP. 
 \paragraph{Math as trees.} Many approaches do not incorporate formula structure. For problems involving multi-variate mathematical terms, obvious choices for this are OPTs and SLTs (Fig.~\ref{tree_repr}). For example, Approach0 considers formula OPTs, \textit{without learning}, to achieve competitive results. Inclusion of OPTs during BERT training has been shown to improve performance over BERT in formula retrieval, formula headline generation, and formula topic classification~\cite{peng2021mathbert}, and is also used in math question answering~\cite{mansouri2021dprl}. 
 \paragraph{Combining complementary representations from the same input.} Combined use of OPTs and SLTs of the same formula has been suggested to improve formula retrieval performance~\cite{davila2017layout,mansouri2019tangent, mansouri2021dprl}. This extends to dual-modality mathematical language input. \citet{shen2020solving} obtain sequence and graph encodings of MWPs, and use sequence and tree-based decoders in unison, with an ablation describing advantage over single encoder representations. The leading MWP solver~\cite{zhang2022multi} generates two independent solution expression embeddings, by top-down decomposition~\cite{xie2019goal} and bottom-up construction, which are projected into the same latent space.
 \paragraph{Conclusion.} Delivering mathematical reasoning over discourse requires close integration between step-wise inference control over localised explicit representations (symbolic perspective), and distributed representations to approximate and cope with incomplete knowledge (neural perspective). The current spectrum of mathematical language processing techniques elicits the key components, representational choices and tasks which are central to the conceptualisation of mathematical inference. Integrating the best-performing representational choices across different subtasks, such as distinct mechanisms for processing natural language and formulae, learning complementary representations of mathematical problem text, and incorporating formula structure, represents a short-term opportunity to develop mathematically robust models capable of more coherent argumentation, reasoning, and retrieval.

\section*{Acknowledgements}

This work was partially funded by the Swiss National Science Foundation (SNSF) project NeuMath (200021\_204617).


\bibliography{tacl2021}
\bibliographystyle{acl_natbib}

\appendix

\section{Approach-specific limitations}

\noindent \textbf{Identifier-definition extraction limitations.} Methods considering the link between identifiers and their definitions have split off into at least three recent tasks: identifier-definition extraction~\cite{schubotz2017evaluating,alexeeva2020mathalign}, variable typing~\cite{stathopoulos2018variable}, and notation auto-suggestion~\cite{jo2021modeling}. A lack of consensus on the framing of the task and data prevents a direct comparison between methods. \citet{schubotz2017evaluating} advise against using their gold standard data for training due to certain extractions being too difficult for automated systems, among other reasons. They also propose future research should focus on recall due to current methods extracting exact definitions for only 1/3 of identifiers, and suggest use of multilingual semantic role labelling~\cite{akbik2016multilingual} and logical deduction~\cite{schubotz2016getting}. Logical deduction is partially tackled by \citet{alexeeva2020mathalign}, which is based on an open-domain causal IE system~\cite{sharp2019eidos} with Odin grammar~\cite{valenzuela2016odin}, where temporal logic is used to obtain intervals referred to by pre-identified time expressions~\cite{sharp2019eidos}. We assume the issues with superscript identifiers (such as Einstein notation \textit{etc.}) from \citet{schubotz2016getting} carry over into \citet{schubotz2017evaluating}. The rule-based approach proposed by \citet{alexeeva2020mathalign} attempts to account for such notation (known as \textit{wildcards} in formula retrieval). They propose future methods should combine grammar with a learning framework, extend rule sets to account for coordinate constructions, and create well-annotated training data using tools such as PDFAlign and others~\cite{asakura2021miogatto}. \newline 

\noindent \textbf{Formula retrieval limitations.} \citet{zhong2019structural} propose supporting query expansion of math synonyms to improve recall, and note that Approach0 does not support wildcard queries. \citet{zhong2020accelerating} later provides basic support for wildcards. Tangent-CFT also does not evaluate on wildcard queries, and the authors suggest extending the test selection to include more diverse formulae, particularly those that are not present as exact matches. They propose integrating nearby text into learned embeddings. MathBERT~\cite{peng2021mathbert} performs such integration, but does not learn n-gram embeddings. MathBERT evaluates on non-wildcard queries only. \newline

\noindent \textbf{Informal premise selection limitations.} Limitations involve a lack of structural consideration of formulae and limited variable typing abilities. \citet{ferreira2020premise} note that the graph-based approach to premise selection as link prediction struggles to encode mathematical statements which are mostly formulae, and suggest inclusion of structural embeddings (\textit{e.g.} MathBERT~\cite{peng2021mathbert}) and training BERT on a mathematical corpus. They also describe value in formulating sophisticated heuristics for navigating the premises graph. Later, following a Siamese network architecture~\cite{ferreira2021star} reliant on dual-layer word/expression self-attention and a BiLSTM (STAR), the authors demonstrate that STAR does not appropriately encode the semantics of variables. They suggest that variable typing and representation are a fundamental component of encoding mathematical statements. \citet{hancontrastive} plan to explore the effect of varying pre-training components, testing zero-shot performance without contrastive fine-tuning, and unsupervised retrieval. \citet{coavoux2021learning} propose a statement-proof matching task akin to informal premise selection, with a solution reliant on a self-attentive encoder and bilinear similarity function. The authors note model confusion due to the proofs introducing new concepts and variables rather than referring to existing concepts. \newline 

\noindent \textbf{Math word problem limitations.} In Graph2Tree-Z~\cite{zhang2020graph}, they suggest considering more complex relations between quantities and language, and introducing heuristics to improve solution expression generation from the tree-based decoder. In EPT, \citet{kim2020point} find error probability related to fragmentation issues increases exponentially with number of unknowns, and propose generalising EPT to other MWP datasets. HGEN~\cite{zhang2022hgen} note three areas of future improvement: Combining models into a unified framework through ensembling multiple encoders (similar to \cite{ferreira2021star}); integrating external knowledge sources (\textit{e.g.} HowNet~\cite{dong2003hownet}, Cilin~\cite{hong2008word}); and real-world dataset development for unsupervised or weakly supervised approaches~\cite{qin2020semantically}. \newline

\noindent \textbf{Informal theorem proving limitations.} \citet{wang2020exploration} suggest the development of high-quality datasets for evaluating translation models, including structural formula representations, and jointly embedding multiple proof assistant libraries to increase formal dataset size. \citet{szegedy2020promising} argues that reasoning systems based on self-driven exploration without informal communication abilities would suffer usage and evaluation difficulties. \citet{wuautoformalization2022} note limitations with text window size and difficulty storing large formal theories with current language models. After proposing the NaturalProofs dataset, \citet{welleck2021naturalproofs} characterize error types for the full-proof generation and next-step suggestion tasks, noting issues with: (1) hallucinated references, meaning the reference does not occur in
NaturalProofs; (2) non-ground-truth reference,
meaning the reference does not occur in the ground-truth proof; (3) undefined terms; and (4) improper or
irrelevant statement, meaning a statement that is
mathematically invalid (\textit{e.g.} $2/3
\in \mathbb{Z}$) or irrelevant
to the proof; and (5) statements that do not follow
logically from the preceding statements. Dealing with research-level physics, \citet{meadows2021similarity} note the significant cost of semi-automated formalisation, requiring detailed expert-level manual intervention. They also call for a set of well-defined computer algebra operations such that robust mathematical exploration can be guided in a goal-based setting.

\section{Diagrammatic categorisation of approaches}

\begin{figure*}[htp!]
    \centering
    \includegraphics[width=1\textwidth]{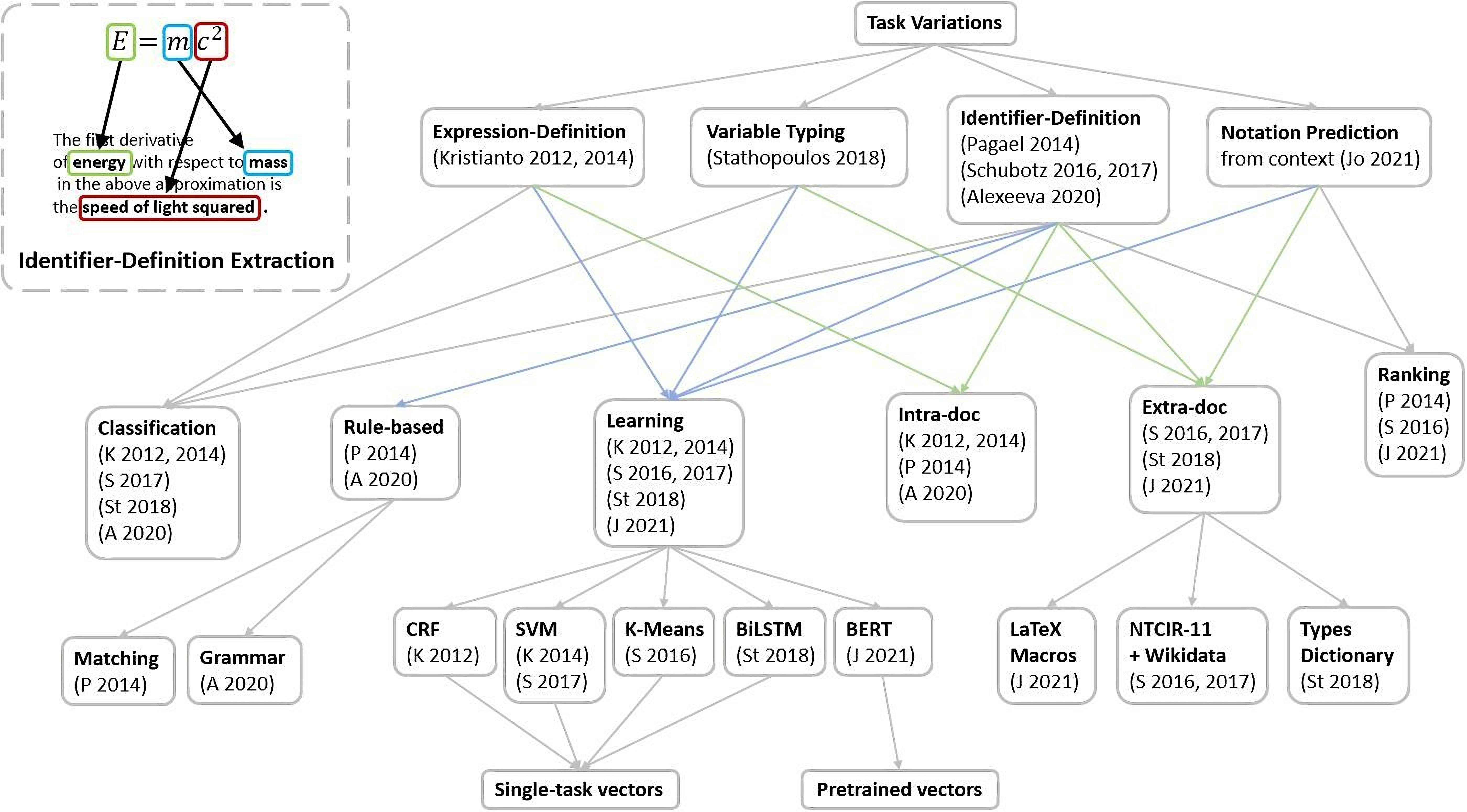}
    \caption{Categorisation of approaches related to identifier-definition extraction. The shorthand notation used such as (K 2012) and (J 2021) refer to the references in the first four boxes, \textit{i.e.}, (Kristianto 2012) and (Jo 2021). The first four boxes are task variations, then arrows point to other categories that may group approaches. For example, (Stathopoulos 2018) is \textit{Variable Typing}, considers a \textit{Classification} task, involves a large machine \textit{Learning} element, uses a \textit{BiLSTM}, learns \textit{Vector} representations of input text without pretraining, relies on information outside of the instance text (\textit{Extra-doc}), which is a \textit{Types Dictionary}.}
    \label{fig:IDE_taxonomy}
\end{figure*}

\begin{figure*}[htp!]
    \centering
    \includegraphics[width=1\textwidth]{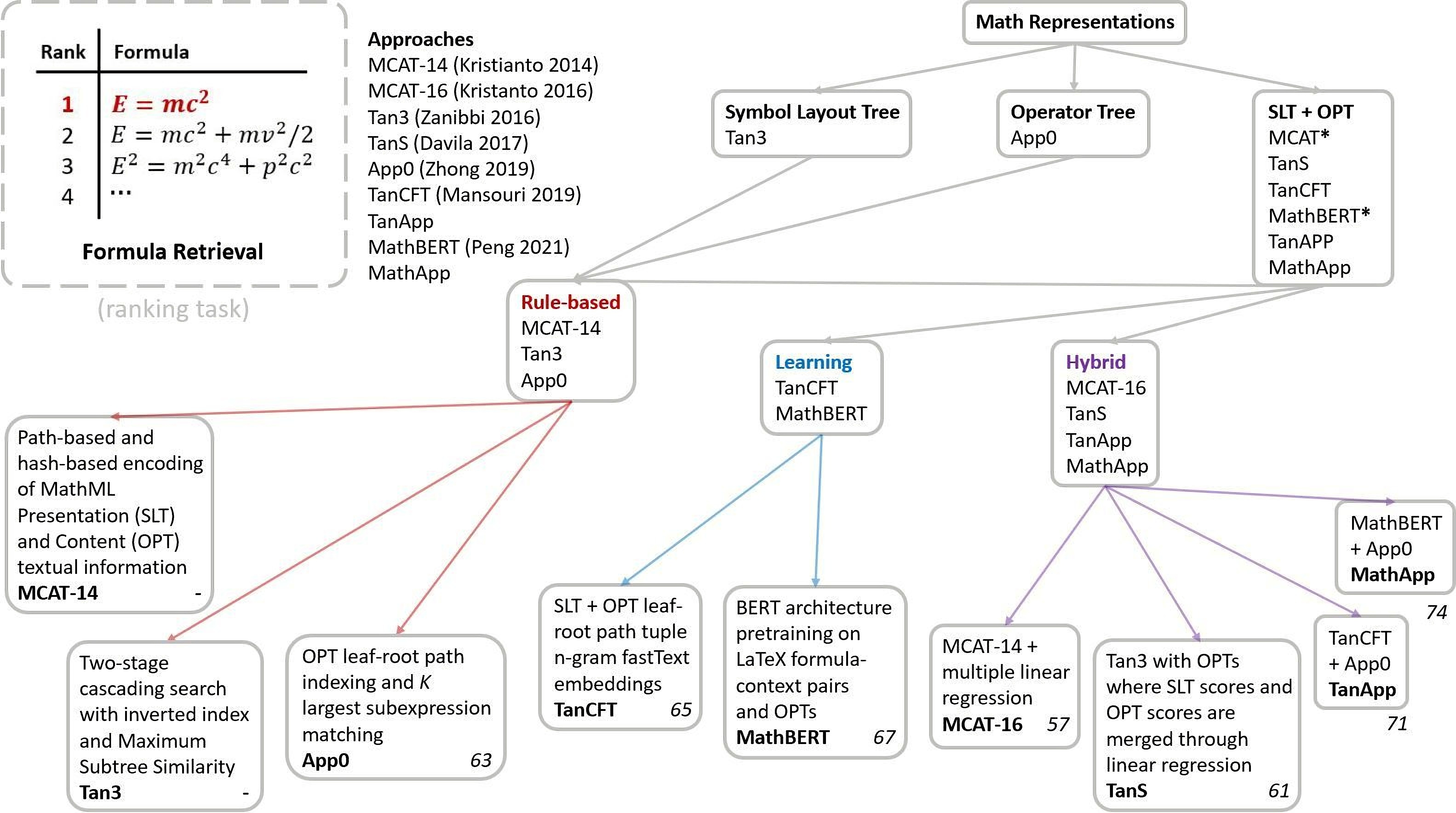}
    \caption{Categorisation of approaches in formula retrieval. The number at the bottom right of boxes refers to their respective Bpref score~\cite{peng2021mathbert}.}
    \label{fig:FR_taxonomy}
\end{figure*}

\begin{figure*}[htp!]
    \centering
    \includegraphics[width=1\textwidth]{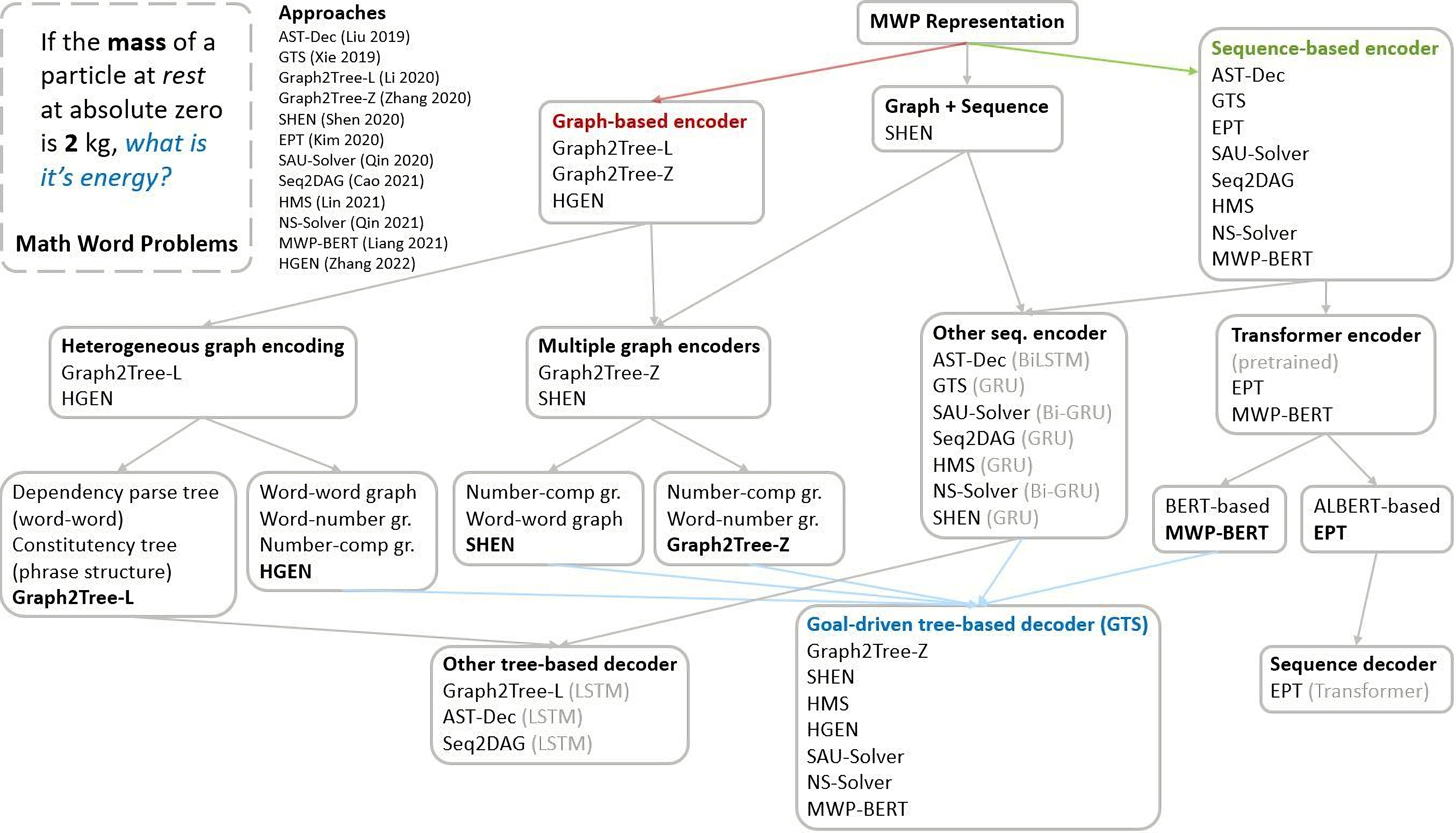}
    \caption{Categorisation of approaches in math word problem solving.}
    \label{fig:MWP_taxonomy}
\end{figure*}

\end{document}